# Machine Learning Suites for Online Toxicity Detection

David Noever
Sr. Technical Fellow, PeopleTec, Inc.
4901-D Corporate Drive
Huntsville, AL 35805
david.noever@peopletec.com

**Abstract**
To identify and classify toxic online commentary, the modern tools of data science transform raw text into key features from which either thresholding or learning algorithms can make predictions for monitoring offensive conversations. We systematically evaluate 62 classifiers representing 19 major algorithmic families against features extracted from the Jigsaw dataset of Wikipedia comments. We compare the classifiers based on statistically significant differences in accuracy and relative execution time. Among these classifiers for identifying toxic comments, tree-based algorithms provide the most transparently explainable rules and rank-order the predictive contribution of each feature. Among 28 features of syntax, sentiment, emotion and outlier word dictionaries, a simple bad word list proves most predictive of offensive commentary.

**Introduction**

In 2015, the Twitter CEO, Dick Costello, took personal responsibility for online harassment, trolling and abuse on the Twitter platform. In an internal company memo (Tiku & Newton, 2015), Costello remarked "We suck at dealing with abuse and trolls on the platform and we've sucked at it for years." In the same company memo, Costello went on to make a business case, "We lose core user after core user by not addressing simple trolling issues that they face every day." To counter such abuse, one naive implementation might enforce censorship keyed to offensive words. Efforts that are more sophisticated might attempt to contextualize speech and capture more subtle tones such as sarcasm ("Your view is just wonderful") which can tilt classification either way depending on one's intonation. Since 2015 when researchers lamented the paucity of published work on toxic speech (Djuric, et al. 2015), many subsequent investigators have sought to recognize abusive language and potentially block cyberbullies, partly because they cause real harm but also because they undercut further online growth.

| Comment | Toxic Rating | |
|---|---|---|
| We suck at dealing with abuse and trolls on the platform and we've sucked at it for years | 0.77 | Toxic |
| We suck at dealing with abuse and trolls on the platform but we'll get better at it. | 0.74 | Toxic |
| We don't suck at dealing with abuse and trolls on the platform and we've never sucked at it | 0.64 | Unsure |
| We're not good at dealing with abuse and trolls on the platform and we've sucked at it for years | 0.61 | Unsure |
| We're not good at dealing with abuse and trolls on the platform but we'll get better at it. | 0.35 | Unlikely |
| We're not good at dealing with fame and fortune on the platform but we'll get better at it. | 0.06 | Unlikely |

*Figure 1 Word-based adversarial attack on Perspective API rating system*

In one recent effort, Google's new machine learning model, called Perspective API, provides a public moderation tool to detect negative behavior (English only). Collaborators on the project included Wikipedia, The New York Times, The Economist, and the Guardian, where the publishing outlets employ the "anti-toxin" algorithm for some first-line filtering of reader comments. The model derives its toxicity scores from a convolutional neural net, or CNN (Conversational AI, 2018). In answer to the key question, "What's toxic?" their API documentation indicates, "This model was trained by asking people [at least 10 per line] to rate internet comments on a scale from 'Very toxic' to 'Very healthy' contribution. Toxic is defined as "a rude, disrespectful, or unreasonable comment that is likely to make you leave a discussion." Their definition is transactional: offense is not a moral principle but derives from the likelihood of exiting a discussion. In a demonstration of the central challenge, we performed an introductory thought experiment: submit to Perspective the Twitter CEO's original comment to find its probable toxic likelihood (in a machine-learning context) and then alter the sentence structure incrementally to lower its severity. As shown in Figure 1, this experimental reconstruction illustrates some language challenges, akin to an adversarial attack on the Perspective API. Like the subtle changes of tone used by Hosseini, et al (2017) and Gröndahl, et al. (2018), such adversarial attacks can reveal brittleness in toxic speech detectors.

One might assume initially that nothing in the CEO's comment would offend, since Costello offered it internally to his company as a *mea culpa* for not ending abuse. A sympathetic human reader might categorize it on the Perspective

API scale, as "Very healthy". Like most deep learning methods, however, the Perspective API provides no obvious path to explain the underlying reasons behind its prediction. Their zero-to-one scale calibrates toxic probability, not severity; the algorithm labels the original Twitter apology as "Toxic" with 0.77 likelihood. We posit that one keyword ("suck") provides some basis for the score, since it resembles patterns in comments that people have tagged as toxic. To probe the influence of raw keyword counts, we find that removing one instance lowers the probability (0.74) but not significantly. By altering the sentence to negate the keyword ("don't suck") and its meaning, we lower the rating to "Unsure" (0.64) and highlight that the algorithm includes enough contextual features to capture word bi-grams (or n-grams). However, just as this algorithm ironically takes offense at Twitter's original apology (for hosting offensive tweets), this negation also lowers the toxicity when the comment now reads as a company boast about how effectively the platform has blocked all offensive behavior. At this point, even a sympathetic human reader might view Twitter as sounding arrogantly over-confident and leave the discussion, thus satisfying the definition of "Toxic". To drain all potential offense from the original comment, one keyword ("suck") and the subject ("abuse and trolls") renders the comment benign. More subtle adversarial experiments might keep the core content, but lengthen the comment via two-fold repetition (Toxic, 0.82), change from plural ("We") to first person ("I") (Toxic, 0.79), or add an exclamation (Toxic, 0.79).

In part to improve its scoring, Jigsaw, Google and Wikipedia released their training and test data to international data competitors (Kaggle.com, 2018). Two goals were to set a new benchmark and to sample alternative solutions, including direct use of the Perspective API. Because both its method (CNN) and data (Jigsaw) are now open, the research community can also investigate deeper criticism of machine-learning bias (Greenberg, 2017), an outcome that potentially proves deeply entrenched when human crowds label (and eventually censor) offensive speech. While some criticism stems rightly from simply learning from and mimicking the pattern of racial, gender and ableism bias found in toxic training data, a core research issue follows from the algorithm's opaque reasoning itself. Explainable rules for machine learning have become a priority with the implementation of Europe's June 2018 General Data Protection Regulation (GDPR).

To explore these issues, we investigate large suites (60+) of machine learning algorithms on the same Jigsaw dataset, with three research goals. Firstly, we evaluate each algorithm's performance and execution times. Secondly, we compare and contrast their transparency (explainable rules). Finally we rank-order the feature importance to highlight their relative predictive value. The latter task may guide the tuning of new features that can counteract some of the adversarial methods illustrated by the original Twitter apology. In a direct example of the challenge, a first step to tokenize words in natural language processing commonly may strip punctuation (like commas), yet inadvertently construct a new and much more offensive meaning ("Time to eat children").

**Methods**
*Jigsaw Data.* As compiled by Jigsaw and Google's Counter-Abuse Technology Team, the Toxic Comment Classification Challenge data captures 159,571 comments from Wikipedia Talk Pages as training data with a similar but unlabeled test data size. As rated by human scorers, the training labels include seven types of comment annotations. The frequency of each label type skews heavily towards a dominant benign rating as shown in parenthesis: benign (89.83%), toxic (3.55%), severe toxic (0.03%), obscene (1.40%), threat (0.10%), insult (4.21%), and identity hate (0.88%). While perhaps consistent with real-world log frequencies, this label imbalance requires some consideration; otherwise, all algorithms could naively achieve 90% accuracy from predicting all comments as benign. A method to balance this distribution therefore is needed, particularly to handle the tiny fraction of comments labeled as severe toxic (41 total) and threat (163). The goal is to produce a learning model to predict the probability for each toxic comment type.

| Category | Feature |
|---|---|
| Syntax | Capital Words |
| | Normalized Capital Words |
| | Exclamation Count |
| | Question Count |
| | Symbolic Characters |
| | Punctuation Count |
| | Word Count |
| | Normalized Word Count |
| | Sentence Count |
| | First Person Count |
| | Plural Pronoun Count |
| Dictionaries | Stop Word Count |
| | Bad Word Count |
| | Fraud Word Count |
| Sentiment | Bing Sentiment |
| | AFINN Sentiment |
| | Negative Sentiment |
| | Positive Sentiment |
| Emotion | Anger |
| | Anticipation |
| | Disgust |
| | Fear |
| | Joy |
| | Trust |
| | Sadness |
| | Surprise |

*Figure 2 Feature generation for Jigsaw metadata extraction*

One approach to data preparation for unbalanced data is to choose metrics that are insensitive to class distribution. The challenge itself scores test success using the average of the individual precision-recall curves (area under curve, AUC) of each predicted class label. This metric reveals the true positives (y) vs. true negatives (x) for each label, and their overall class average thus penalizes all mislabeling irrespective of their total frequency in the dataset. The present work transforms the Jigsaw data into two parallel but balanced sets.

By random under-sampling of over-represented classes, we test many algorithms on the smallest possible balanced training set (41 instances each). This method allows for rapid assessments across many parameters and machine learning algorithms. Under-sampling suffers from information loss among the majority classes, but given the diversity of benign commentary in the wild, one may tolerate some reduced representation when training. This small dataset also allows quick evaluation of which features might prove uncorrelated and predictive for toxic classes. Following this broad algorithm survey, we leave the majority class (benign) unchanged and over-sample the other minority classes to enlarge the original 159,571 cases to over a million examples (7 classes balanced with 143,346 instance each). Over-sampling does not produce information loss, but replication of the minority classes can over-fit and perform less accurately against unseen test data.

*Feature Generation.* Because we are mainly interested in rank-ordering features and explaining decision rules, we chose to simplify the menu of predictive factors to four broad families: syntax, dictionaries, sentiment and emotion. As shown in Figure 2, we compiled 28 basic features in total from these four families. Figure 3 shows the correlation between the features and reflect their familial relationships, particularly the correlations between emotional factors.

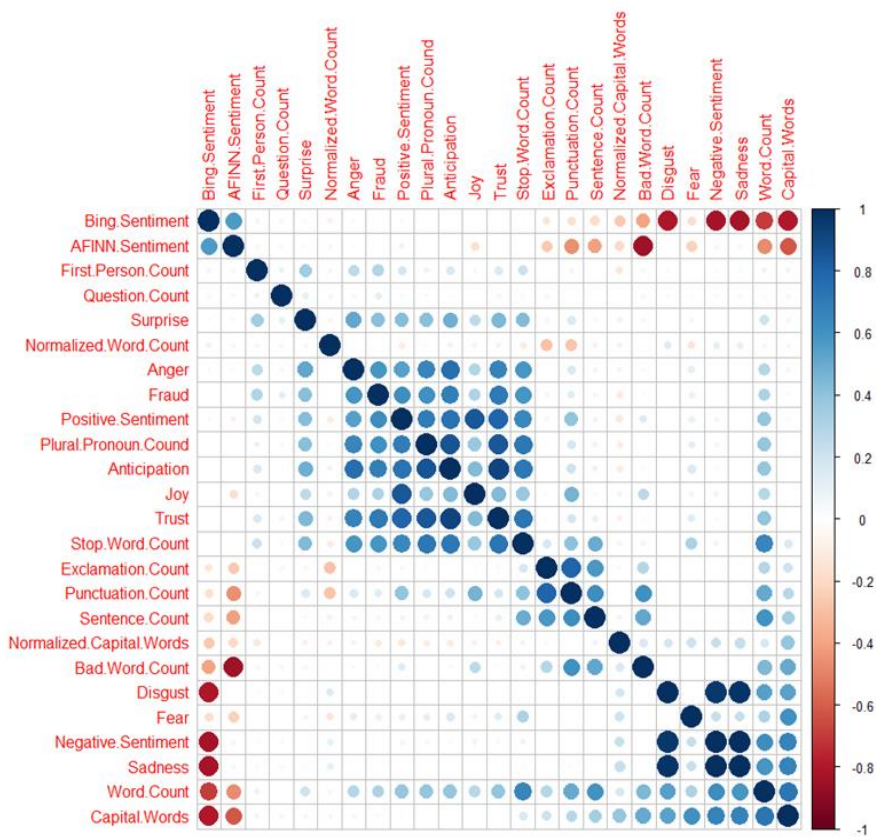

*Figure 3 Correlation matrix of toxic features*

We analyzed sentiment factors in comments in a multistep process. We accounted for single word stems by first counting, then removing numbers, stripping punctuation and lowercasing the text (Mohammad, 2018). For English sentiment, we employed the AFINN-111 scale for positive or negative polarity (Nielsen, 2011). Examples of strong negative sentiment include most curse words, racial or class epithets and grievances (e.g. "tortures"). Examples of strong positive sentiment include broad approvals (e.g. "breakthrough" or "outstanding"). We also apply alternative crowd-sourced dictionaries such as Bing (Liu, 2012) and the NRC-Canada Word-Emotion Association Lexicon (Mohammed & Turney, 2013) with eight basic emotions (anger, fear, anticipation, trust, surprise, sadness, joy and disgust). We applied customized counters for syntactic or character-level features of interest, including: comment length, word count, capitalized letters, punctuation count, such as question marks, exclamation and symbolic non-ASCII characters. We also tabulated the pronoun usage, both first person and plural. Finally, we applied various standardized neutral (stop-words, Fox, 1989) and negative word dictionaries, such as Google's profanity list (see e.g., Squire, et al., 2015) and financial fraud related terms (Goel, et al., 2010) used to predict deceptive writing. It is worth noting that these features highlight single tokenized text and by definition take little or no account of a comment's full context, as one might for instance find in a full n-gram (Brown, 1992) or term frequency matrix (TF-IDF or term-frequency inverse-document frequency, Salton, et al., 1983). This choice, while limiting to overall accuracy, does enable more model transparency about exactly why tree-based algorithms may make a negative prediction.

*Feature Importance.* We applied automated feature selection to rank order the importance of each factor. The approach assigns a top-down search based on the Boruta algorithm (Murin, et al., 2010), which we applied independently to both the under-sampled and over-sampled datasets. Boruta uses the random forest algorithm to search for relevant features and iteratively compares each attribute's importance with achievable random inputs. This method gives a discriminating power for each factor that contributes uniquely to a better prediction. In Figure 4, the most important detection factors for toxic commentary involve bad word dictionaries, punctuation counts, first person pronouns, and capitalized words.

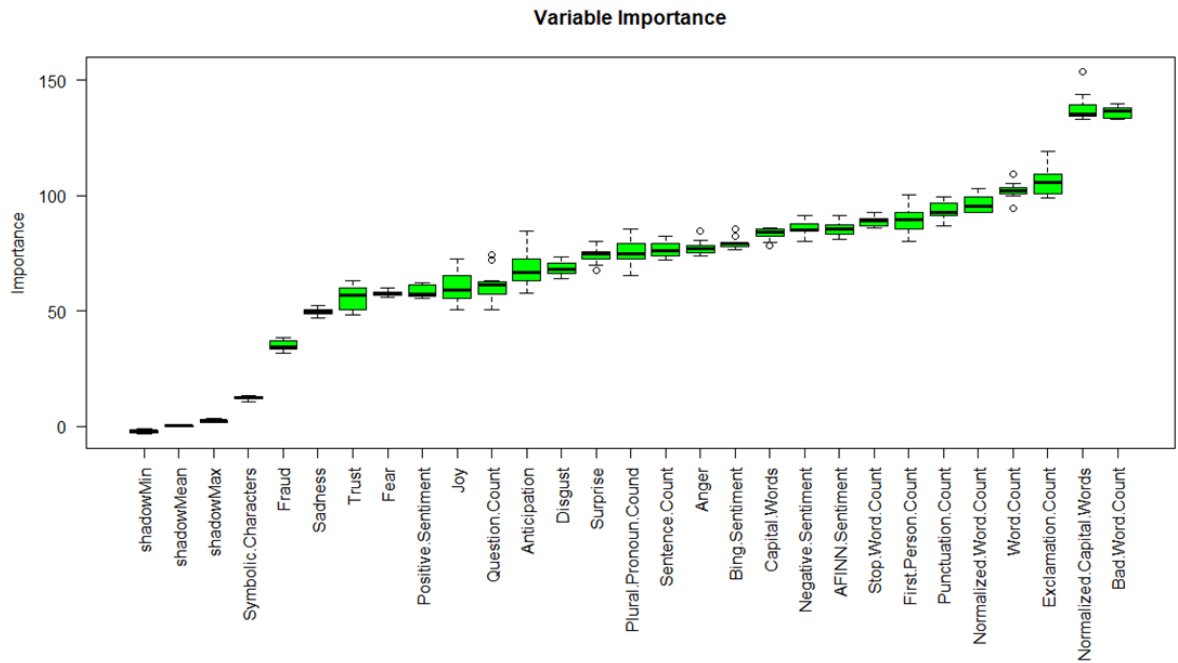

*Figure 4 Rank order for over-sampled dataset and factor importance. The algorithm confirms all 28 features as important, meaning each offers an uncorrelated but predictive input.*

Consistent with Figure 4 visual inspection, the least important factors are emotional features related to sadness, trust, fear, disgust and anger. This suggests that looking for unhealthy comments in the Jigsaw dataset may not classify toxic behavior using simple emotionally charged keywords. The broad contribution of syntactic features like capitalized words and punctuation suggests a multi-factor prediction algorithm might outperform single-factor filters for designing online toxicity monitors.

*Classifier Suites.* We implemented the classifier suite using the R algorithm libraries conveniently packaged in a larger wrapper library called caret (Kuhn, 2008). This machine-learning wrapper provides a common calling function, "train", which simplifies large-scale algorithm surveys. For toxic comment detection, the classifier suites include major families of linear multinomial and adaptive regressions; discriminant and principal component analysis; Bayesian; neural networks including deep and shallow layers; support vector machines; rule-based decision trees include ensembles such as boosting, bagging, stacking and random forests. We rank the algorithms' predictive accuracy and relative execution time. We validate performance using training and testing data partitioning, along with repeated cross-fold validation. We attempted no parameter tuning or rescaling in favor of accepting model defaults. We excluded models if they took longer than a few hours or failed to converge. Because the Jigsaw multi-classes are categorical, about two-thirds of the 189 total (caret) algorithms prove either inapplicable, crash or fail to converge.

Rather than dwell on the particular aspects of these non-starters, we rank the 62 algorithms for performance and relative execution time. Performance on the re-balanced datasets can be simply interpreted as the percent correct (overall accuracy) across all seven classes. We re-scale the relative execution times between zero and one, where one corresponds to the slowest algorithm to complete.

**Results**

We summarize the broad algorithm survey both by its specific implementation (Figure 5) and by the algorithmic family (e.g. boosted tree family, Figure 6). These results for 62 algorithms show that the model-averaged neural network achieves relatively moderate accuracy (40%) in average relative time (0.5). In Figure 7, we list the specific R algorithm used in multiple bar graphs showing the accuracy, Kappa and log time for execution. Kappa here takes additional account of random choices and thus highlights the algorithm's discriminating power over chance selection (Cohen, 1960). While the overall accuracy is low, the initial objective using the under-sampled (41) instances of each toxic class focuses on relative performance to prioritize future runs with the larger (one million comments) dataset.

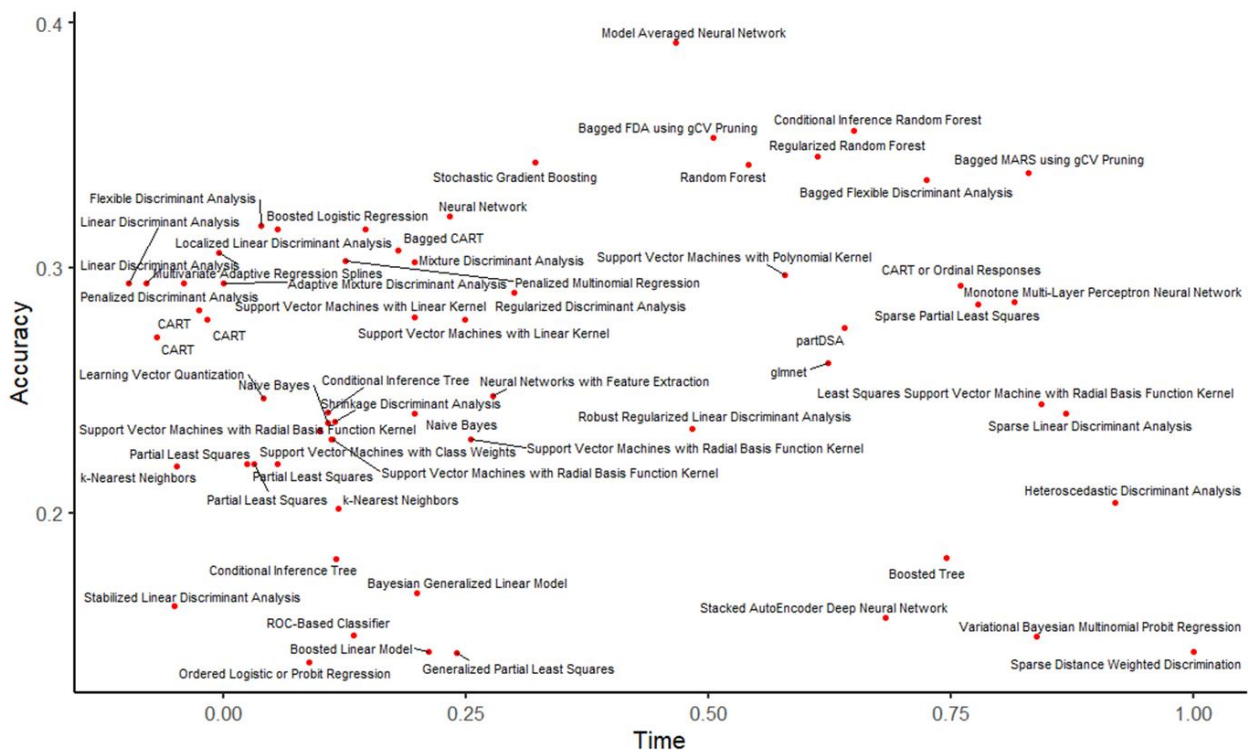

*Figure 5 Algorithm Accuracy vs. Time for Predicting Multi-class Toxicity*

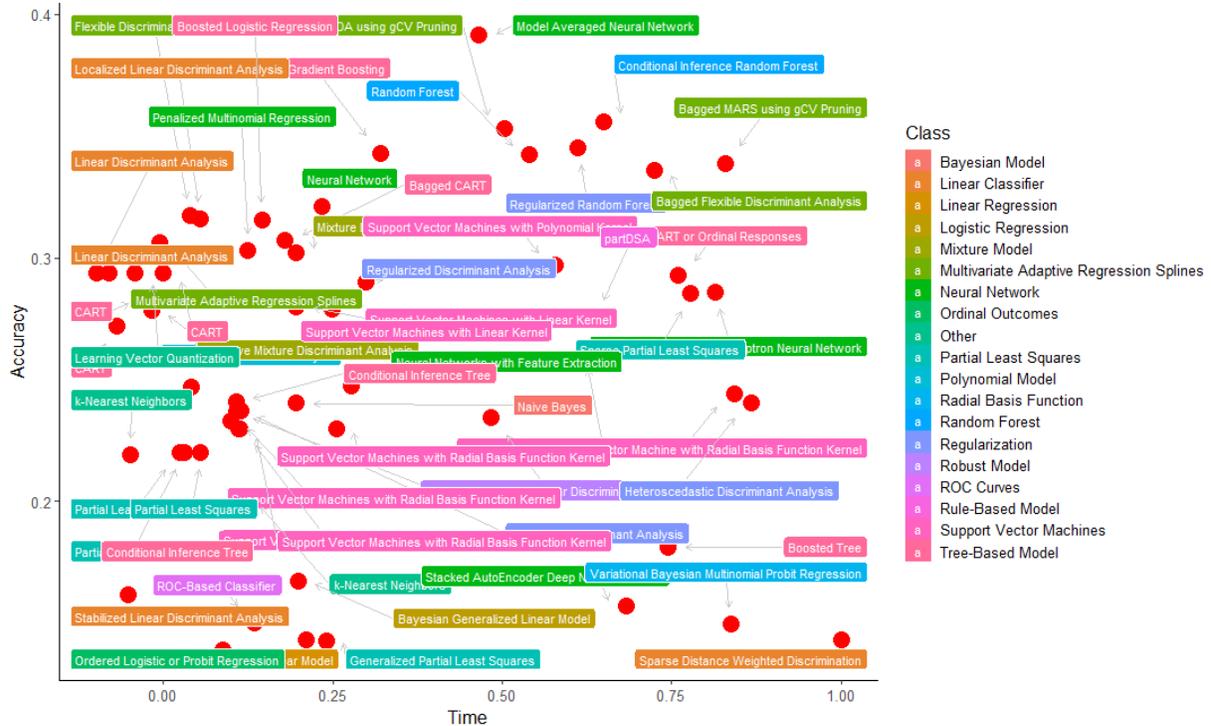

*Figure 6 Algorithm performance color-coded by family*

To understand how the algorithms might adjudicate a negative comment, we extracted rules from a simple decision tree (CART). Such single-tree models are well-known to over-fit training data and perform inflexibly when confronted with unseen testing data. In Figure 8, these rules offer a deeper "if-then" ruleset than just applying censor dictionaries might; they also capture some subtle differences between the multi-classes and complex decision-making: in other words, how might a machine differentiate "Toxic" from "Obscene" or "Identity-Hate". To trace the decision about hate speech (class 6, or lower right box and green label in the Figure 8), for example, requires the comment to include a relatively high bad word count, less emotion of anticipation than the other toxic speech categories and a relatively higher count of common stop words. Other similar classes like "Insult" (class 5 in Figure 8) may share a similar trunk but branch out if the comment also features more capital letters ("textual screaming") and punctuation. The goal here remains not so much to establish blanket rules but to discover underlying factors that different flavors of negativity may manifest. Such an exercise can guide both new factors to explore but also simplify the often brute-force or blind ensemble of many useless filtering methods.

**Relation to Previous Work**
Parekh, et al. (2017) reviewed multiple algorithms for detecting toxic commentary from 2009 to the present; they note for the binary classification identifying "toxic vs. nontoxic", the diversity of approaches spanning linear regression, support vector machine, and Naïve Bayes. A typical case involves introducing a new feature model and applying a few algorithms to compare their performance. Even when different papers use the same dataset, their feature choices make direct algorithmic comparisons difficult. Pavlopoulos, et al. (2017) examined deep learning methods with word embeddings on a variety of online commentaries (both Greek and English Wikipedia). They found high 95+% Area-Under-Curve metrics. In contrast the present algorithm comparison does not deploy word embeddings and the results for generic deep learning settle on the low end of accuracy (<20%) and offer longer execution times than median algorithms (e.g., see caret defaults for R package deepnet, Rong, 2014, which include 3 hidden layers only). The points marked in Figures 5-6 for "Stacked Autoencoder Deep Neural Network" and in Figure 7 for "dnn" indicate its lower performance. Given the much larger data requirements for training many successful deep learning methods, such underperformance is partially expected on fewer than 200,000 comments and an unbalanced dataset. While not state-of-the-art as used here, we postulate that: 1) the feature selection steps for word encoding explain the greater performance seen by Pavlopoulos, et al, as much as the algorithm choices; 2) text as input to a recurrent neural network (RNN) or other deep architectures (e.g. convolutional nets) needs more attention to training and feature selection compared to traditional image recognition problems.

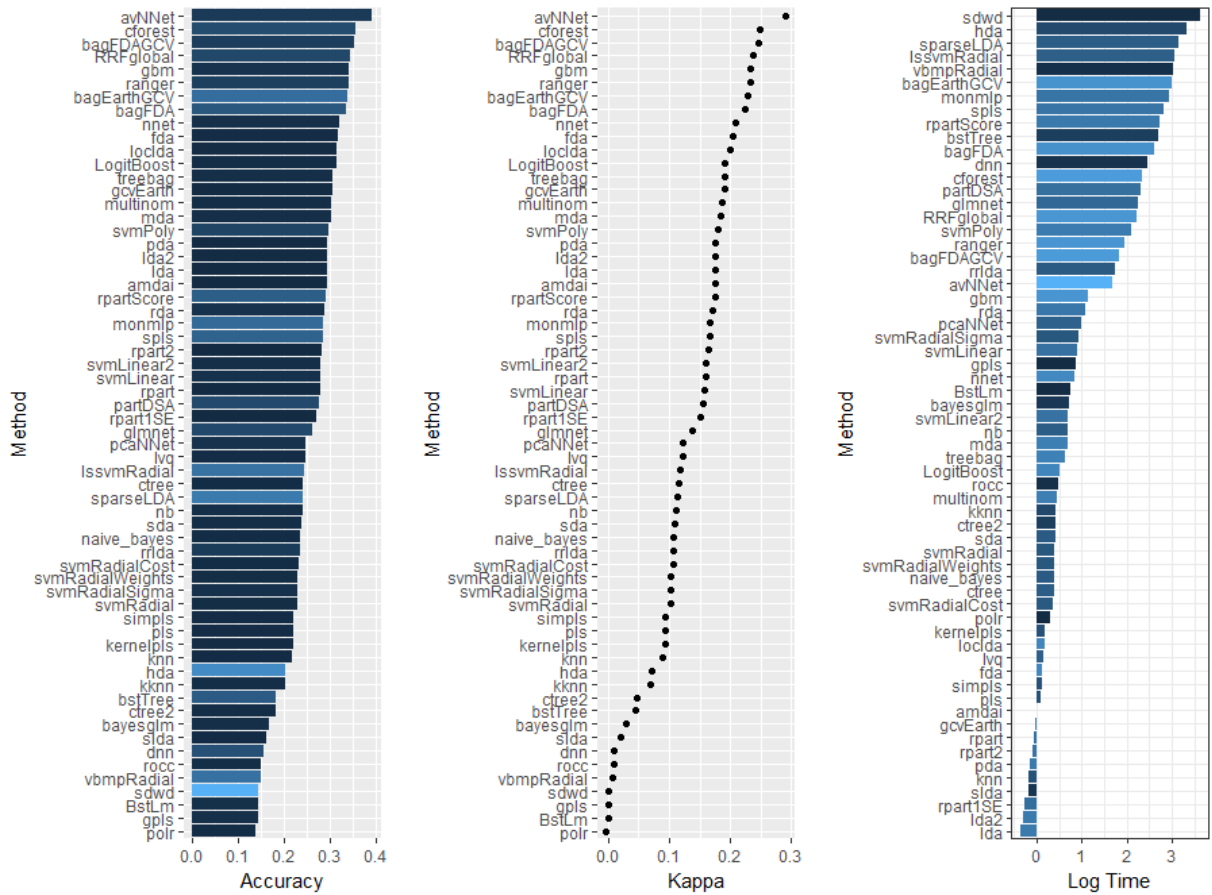

*Figure 7 Bar chart of algorithm accuracy, Kappa and execution time by specific R library algorithm name*

One intriguing finding of many data competitions centers on the ability of many weak learners (ensembles) to outperform a single strong learner. This pooled voting and ensemble method dates most notably back to the original Netflix Million Dollar competition where data teams eventually won with combinations of over 100 algorithms (Bell, et al., 2008). One might assume that each weak algorithm is really just isolating a feature subset that another sufficiently advanced or master algorithm would be better suited to discover. For example, in image analysis, the depth and design of the deep learning network itself all but removes feature engineering, since raw unprocessed pixel values get systematically filtered into new features through each layer. The architecture becomes the feature. In ensembles of weak learners, the votes of our algorithm survey can be assessed using the now closed Kaggle competition. Using a random forest trained on just the 41 under-sampled data points, we generalize to the 157,000 unseen test cases with an overall score of 0.5782. However, if we combine this overall best algorithm with some mixture of under-performing ones (RPART, SVM and GLM), we raise the score by 5%. While neither the raw score, nor the incremental improvement by combining algorithms, is outstanding, the observation that only 41 examples in a random forest can achieve any success at all is notable, along with what ensembles add that amounts to more than just averaging down the overall score.

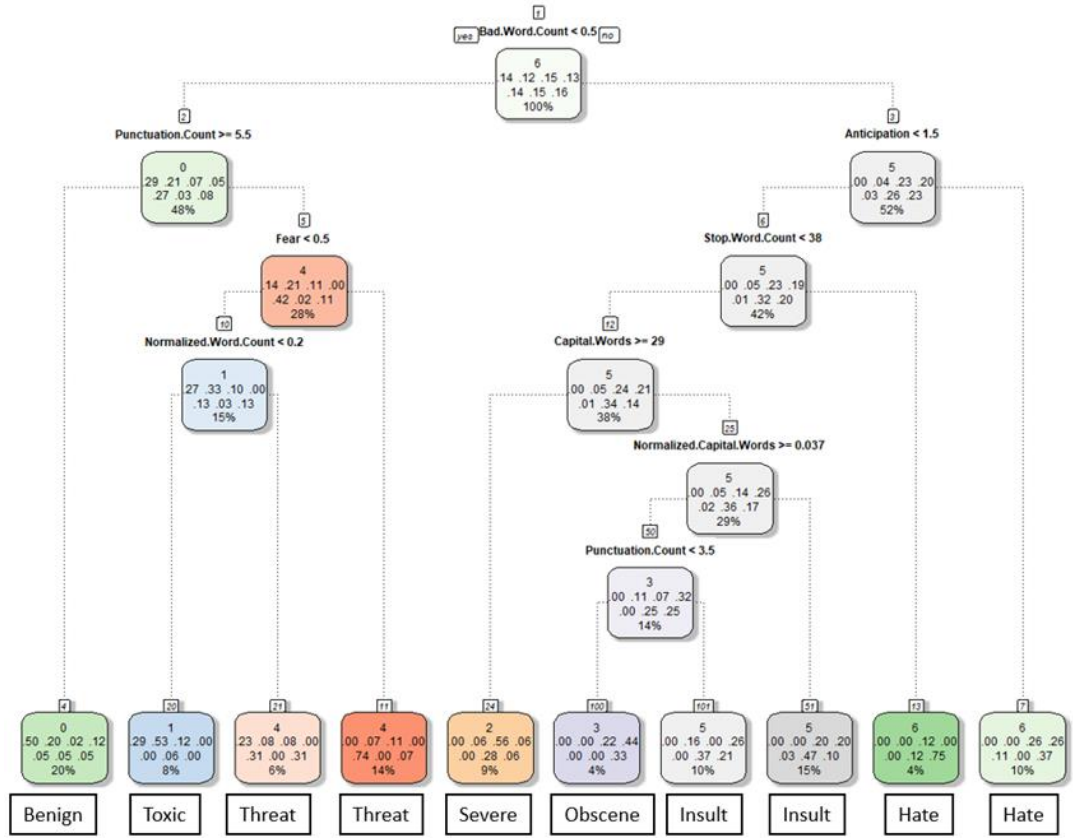

*Figure 8 Decision Tree Rule-set for Toxic Comment Classification*

**Conclusions and Future Work**

We systematically evaluated 62 classifiers representing 19 major algorithmic families against features extracted from the Jigsaw dataset of Wikipedia comments. We compared the classifiers based on statistically significant differences in accuracy and relative execution time. Tree-based algorithms provided both the most transparently explainable rules and a rank-order list for the most important factors. Among 28 features of syntax, sentiment, emotion and outlier word dictionaries, a simple bad word list proves most predictive of offensive commentary. When holding the feature set constant but applying a broad and varying suite of algorithms, the fair comparison offers insight into the complexities of natural language parsing and identifies promising algorithms to include in future detection ensembles. A working hypothesis worth testing in this framework is whether certain algorithm classes are definable as factors themselves (e.g. votes) in a much bigger and more effective detection workflow (e.g. sum of votes). We anticipate testing this proposition quantitatively in future work and other methods to make toxic detection less brittle or easily evaded (Gröndahl, et al. 2018).

**Acknowledgements**

The author would like to thank the PeopleTec Technical Fellows program for encouragement and project assistance.